%% file: bare_conf.tex
\documentclass[conference]{IEEEtran}
\usepackage{cite}
\usepackage{amsmath,amssymb,amsfonts}
\usepackage{amsthm}
\usepackage{algorithmic}
\usepackage{acronym}
\usepackage{textcomp}
\usepackage{subcaption}
\usepackage{float}
\usepackage{array}
\usepackage{tabularx}
\usepackage{adjustbox}
\usepackage{soul, color, acronym, multirow, balance}
\usepackage[]{graphicx}
\usepackage{epsfig, epstopdf}
\usepackage{booktabs}
\usepackage[hyphens]{url}
\usepackage[justification=centering]{caption}
\usepackage[colorinlistoftodos]{todonotes}

\usepackage[authormarkup=superscript,todonotes={textsize=tiny}]{changes}
\definechangesauthor[name={Lorenzo}, color=red]{LV}
\definechangesauthor[name={Pietro}, color=green]{PC}
\definechangesauthor[name={Alberto}, color=blue]{AG}
\definechangesauthor[name={All}, color=orange]{ALL}
\definecolor{orcidlogocol}{HTML}{A6CE39}
\usepackage{soul}

\theoremstyle{plain}
\newtheorem*{Def}{Definition}

\usepackage{enumitem}

\input{sections/acr.tex}

\begin{document}
\title{Federated Semi-Supervised Classification of Multimedia Flows for 3D Networks}
\author{
		\IEEEauthorblockN{Saira Bano\IEEEauthorrefmark{1}\IEEEauthorrefmark{2}, Achilles Machumilane\IEEEauthorrefmark{1}\IEEEauthorrefmark{2}, Lorenzo Valerio\IEEEauthorrefmark{3}, Pietro Cassar\`a\IEEEauthorrefmark{2}, Alberto Gotta\IEEEauthorrefmark{2}}
		\IEEEauthorrefmark{1}Department of Information Engineering, University of Pisa, Pisa, Italy\\
		\IEEEauthorrefmark{3}CNR, Institute of Informatics and Telematics, Pisa, Italy\\
		\IEEEauthorrefmark{2}CNR, Institute of Information Science and Technologies, Pisa, Italy\\ 

	}
\maketitle

\begin{abstract} Automatic traffic classification is increasingly becoming important in traffic engineering, as the current trend of encrypting transport information (e.g., behind HTTP-encrypted tunnels) prevents intermediate nodes from accessing end-to-end packet headers. However, this information is crucial for traffic shaping, network slicing, and \ac{QoS}
management, for preventing network intrusion, and for anomaly detection. 3D networks offer multiple routes that can guarantee different levels of \ac{QoS}. Therefore, service classification and separation are essential to guarantee the required QoS level to each traffic sub-flow through the appropriate network trunk.
In this paper, a federated feature selection and feature reduction learning scheme is proposed to classify network traffic in a semi-supervised cooperative manner. The federated gateways of 3D network help to enhance the global knowledge of network traffic to improve the accuracy of anomaly and intrusion detection and service identification of a new traffic flow.
\end{abstract}

\IEEEpeerreviewmaketitle

\input{sections/Introduction.tex}

\input{sections/RelatedWorks.tex}
\input{sections/ProblemDefinition.tex}

\input{sections/SystemArchitecture.tex}
\input{sections/PerformanceEvaluation.tex}

\input{sections/FeaturesAnalysis.tex}
\section{Conclusion}
\label{sec:conclusions}
In this work, we have developed a procedure to capture the rate of QUIC traffic within a data stream using a regression model trained with a semi-supervised technique. 
This technique is used to intercept and separate traffic belonging to a particular protocol (e.g., QUIC) or class of service (streaming, real-time interactive, browsing, mail, etc.) in order to direct it to a desired trunk of a 3D network. To this end, multiple edge gateways are assumed to cooperate to create a common classification model to intercept and classify the destination traffic.
We tested the effects of federation of regression models trained locally by a set of gateway nodes. Finally, together with the federation, we introduced a novel feature selection technique to reduce communication and computation costs. Based on the cross-entropy method, the feature selection algorithm maximizes the mutual information of the selected features and class attributes, i.e., the quote of QUIC traffic. The proposed cross-entropy algorithm was used and compared in a centralized and a federated distributed form. The selected features provide the same performance as the whole set with negligible RMSE error, while outperforming the other baselines of 5-10\% in the centralized/supervised approach. In addition, cross-entropy was also tested in a federated manner and using a semi-supervised approach with local soft labeling. Such a distributed approach significantly reduced the traffic overhead required to train the regression model, but at the cost of a higher RMSE than the centralized baseline, but preserving the number of rounds of communication required to federate the models.
\section*{Acknowledgment}
This work has been partially supported by ESA SatNEx V project contract n. 4000130962/20/NL/NL/FE, and by TEACHING H2020 projects (GA  \#871385). 
\IEEEtriggeratref{0}
\balance
\bibliographystyle{IEEEtran}
\bibliography{references}
\end{document}

%% file: sections/acr.tex
\acrodef{ASP}{Associated Stochastic Problem}
\acrodef{MI}{Mutual Information}
\acrodef{EA}{Essential Attribute}
\acrodef{FFS}{Federated Feature Selection}
\acrodef{FS}{Feature Selection}
\acrodef{PER}{Packet Erasure Rate}
\acrodef{DPI}{Deep Packet Inspection}
\acrodef{ML}{Machine Learning}
\acrodef{DL}{Deep Learning}
\acrodef{BoF}{Bag-of-Flows}
\acrodef{FL}{Federated Learning}
\acrodef{PLR}{Packet Loss Rate}
\acrodef{PRR}{Packet Received Ratio}
\acrodef{BER}{Bit Error Rate}
\acrodef{SG}{Smart Gateway}
\acrodef{SS}{Smart Switch}
\acrodef{RTT}{Round-Trip Time}
\acrodef{TCP}{Transmission Control Protocol}
\acrodef{UDP}{User Datagram Protocol}
\acrodef{AIMD}{Additive Increase Multiplicative Decrease}
\acrodef{PEP}{Performance Enhancing Proxy}
\acrodef{cwnd}{congestion window}
\acrodef{STP}{Satellite Transport Protocol}
\acrodef{BDP}{Bandwidth-Delay Product}
\acrodef{QoS}{Quality of Service}
\acrodef{LoS}{Line of Sight}
\acrodef{NLoS}{Non Line of Sight}
\acrodef{BLoS}{Beyond Line of Sight}
\acrodef{QoE}{Quality of Experience}
\acrodef{ACM}{Adaptive Coding and Modulation}
\acrodef{DAMA}{Demand Assignment Multiple Access}
\acrodef{VANET}{Vehicular Ad-Hoc Network}
\acrodef{RA}{Random Access}
\acrodef{CRA}{Contention Resolution ALOHA}
\acrodef{SA}{Slotted ALOHA}
\acrodef{DSA}{Diversity Slotted ALOHA}
\acrodef{OBU}{On-Board Unit}
\acrodef{CRDSA}{Contention Resolution Diversity Slotted ALOHA}
\acrodef{SIC}{Successive Interference Cancellation}
\acrodef{ARQ}{Automatic Repeat reQuest}
\acrodef{SC-ARQ}{Selective-Coded ARQ}
\acrodef{SR-ARQ}{Selective-Repeat ARQ}
\acrodef{IRSA}{Irregular Repetition Slotted ALOHA}
\acrodef{CGC}{Complementary Ground Component}
\acrodef{GCS}{Ground Control Station}
\acrodef{RSU}{Road Side Unit}
\acrodef{ACK}{Acknowledgment}
\acrodef{NACK}{Negative Acknowledgment}
\acrodef{DVB-SH}{Digital Video Broadcasting - Satellite Services to Handhelds}
\acrodef{DVB-H}{Digital Video Broadcasting - Handheld}
\acrodef{DVB-RCS2}{Digital Video Broadcasting - Return Channel via Satellite, II generation}
\acrodef{SACK}{Selective Acknowledgment}
\acrodef{SNACK}{Selective Negative Acknowledgment}\acrodef{SNACK}{Selective Negative Acknowledgment}
\acrodef{SNIR}{Signal to Noise plus Interference Ratio}
\acrodef{SCPS-TP}{Space Communications Protocol Specifications - Transport Protocol}
\acrodef{CCSDS}{Consultative Committee for Space Data Systems}
\acrodef{ESA}{European Space Agency}
\acrodef{NASA}{National Aeronautics and Space Administration}
\acrodef{BSM}{Broadband Satellite Multimedia}
\acrodef{RLNC}{Random Linear Network Coding}
\acrodef{NC}{Network Coding}
\acrodef{FIFO}{First In, First Out}
\acrodef{FCFS}{First Come, First Served}
\acrodef{BLER}{Block Error Rate}
\acrodef{GEO}{geosynchronous}
\acrodef{PMF}{Probability Mass Function}
\acrodef{LEO}{Low Earth Orbit}
\acrodef{FTP}{File Transfer Protocol}
\acrodef{CRC}{Cyclic Redundancy Check}
\acrodef{MAC}{Media Access Control}
\acrodef{PHY}{Physical layer}
\acrodef{HTTP}{Hypertext Transfer Protocol}
\acrodef{ISP}{Internet Service Provider}
\acrodef{MSS}{Maximum Segment Size}
\acrodef{BIC}{Binary Increase Congestion control}
\acrodef{AQM}{Active Queue Management}
\acrodef{XCP}{eXplicit Control Protocol}
\acrodef{ECN}{Explicit Congestion Notification}
\acrodef{CA}{Congestion Avoidance}
\acrodef{RED}{Random Early Detection}
\acrodef{TD}{Triple-Duplicate}
\acrodef{TO}{TimeOut}
\acrodef{IP}{Internet Protocol}
\acrodef{WMN}{Wireless Mesh Network}
\acrodef{ssthresh}{Slow-Start threshold}
\acrodef{MPE-IFEC}{Multi Protocol Encapsulation - Inter-burst Forward Error Correction}
\acrodef{FEC}{Forward Error Correction}
\acrodef{CRC}{Cyclic Redundancy Check}
\acrodef{P2P}{Peer-to-Peer}
\acrodef{FMT}{Fade Mitigation Technique}
\acrodef{SGD}{Smart Gateway Diversity}
\acrodef{NCC}{Network Control Centre}
\acrodef{ModCod}{Modulation and Coding}
\acrodef{FIFO}{First-In-First-Out}
\acrodef{WRR}{Weighted Round Robin}
\acrodef{WFQ}{Weighted Fair Queuing}
\acrodef{NS}{Network Simulator}
\acrodef{GSE}{Generic Stream Encapsulation}
\acrodef{PDF}{Probability Density Function}
\acrodef{CDF}{Cumulative Density Function}
\acrodef{CoV}{Coefficient of Variation}
\acrodef{MSC}{Message Sequence Chart}
\acrodef{ESA}{European Space Agency}
\acrodef{LIU}{Lebanese International University}
\acrodef{TUM}{Technical University of Munich}
\acrodef{MSCE-CS}{Master of Science in Communications Engineering - Communications Systems}
\acrodef{DLR}{German Aerospace Center}
\acrodef{NOS}{Network Operating System}
\acrodef{NFs}{Network Functionalities}
\acrodef{NC-SGD}{Network Coding for SGD}
\acrodef{ATSP}{Advanced Transport Satellite Protocol}
\acrodef{STP}{Satellite Transport Protocol}
\acrodef{WMN}{Wireless Mesh Network}
\acrodef{SNR}{Signal-to-Noise Ratio}
\acrodef{SINR}{Signal-to-Interference-plus-Noise Ratio}
\acrodef{LMS}{Land Mobile Satellite}
\acrodef{LTE}{Long-Term Evolution}
\acrodef{M2M}{machine-to-machine}
\acrodef{IoT}{Internet of Things}
\acrodef{IoRT}{Internet of Remote Things}
\acrodef{RA}{Random Access}
\acrodef{UAV}{Unmanned Aerial Vehicle}
\acrodef{UTV}{Unmanned Terrestrial Vehicle}
\acrodef{UAS}{Unmanned Aerial System}
\acrodef{FANET}{Flying Ad-Hoc Network}
\acrodef{MANET}{Mobile Ad-Hoc Network}
\acrodef{VANET}{Vehicle Ad-Hoc Network}
\acrodef{C2}{Command and Control}
\acrodef{DTN}{Delay Tolerant Network}
\acrodef{COTS}{Commercial Off-the-Shelf}
\acrodef{IETF}{Internet Engineering Task Force}
\acrodef{CoAP}{Constrained Application Protocol}
\acrodef{MQTT}{Message Queuing Telemetry Transport}
\acrodef{URI}{Uniform Resource Identifier}
\acrodef{PUB/SUB}{Publish / Subscribe}
\acrodef{RCST}{Return Channel Satellite Terminal}
\acrodef{TDMA}{Time Division Multiple Access}
\acrodef{FDMA}{Frequency Division Multiple Access}
\acrodef{TCDMA}{Turbo Code Division Multiple Access}
\acrodef{PDMA}{Power Division Multiple Access}
\acrodef{WSN}{Wireless Sensor Network}
\acrodef{REST}{Representational State Transfer}
\acrodef{EDGE}{Enhanced Data rates for GSM Evolution}
\acrodef{UMTS}{Universal Mobile Telecommunications System}
\acrodef{LTE}{Long-Term Evolution}
\acrodef{E2E}{End-to-End}
\acrodef{3WHS}{Three-way Handshake}
\acrodef{SCADA}{Supervisory Control And Data Acquisition}
\acrodef{SOA}{Service-Oriented Architecture}
\acrodef{WebRTC}{Web Real-Time Communications}
\acrodef{fps}{frames per second}
\acrodef{SSIM}{Structural SIMilarity}
\acrodef{PSNR}{Peak signal-to-noise ratio}
\acrodef{RPi}{Raspberry Pi}
\acrodef{NFV}{Network Function Virtualization}
\acrodef{OPEX}{Operating Expenditures}
\acrodef{CAPEX}{Capital Expenditures}
\acrodef{MIMO}{multiple-input and multiple-output}
\acrodef{SDN}{Software Defined Network}
\acrodef{VNF}{Virtual Network Function}
\acrodef{MEC}{Mobile Edge Computing}
\acrodef{MTC}{Machine-type Communications}
\acrodef{mMTC}{Massive Machine-type Communications}
\acrodef{HTC}{Human-type Communications}
\acrodef{D2D}{device to device}
\acrodef{IIoT}{industrial IoT}
\acrodef{LPWAN}{Low-Power Wide-Area Network}
\acrodef{CPS}{Cyber-Physical System}
\acrodef{ICT}{Information and Communication Technologies}
\acrodef{SDR}{Software Defined Radio}
\acrodef{GPS}{Global Positioning System}
\acrodef{H2M}{human-to-machine}
\acrodef{MC}{Machine}
\acrodef{LA}{Local Area}
\acrodef{WA}{Wide Area}
\acrodef{HAP}{High Altitude Platform}
\acrodef{LAP}{Low Altitude Platform}
\acrodef{RAN}{Radio Access Network}
\acrodef{EIRP}{Equivalent Isotropically Radiated Power}
\acrodef{LT}{Logistics and Transportations}
\acrodef{eNB}{Evolved Node B}
\acrodef{CDN}{Content Distribution Network}
\acrodef{ITU}{International Telecommunication Union}
\acrodef{SIN}{Space Information Network}
\acrodef{DTA}{Delay Tolerant Application}
\acrodef{DSA}{Delay Sensitive Application}
\acrodef{DTN}{Delay Tolerant Networking}
\acrodef{ISL}{Inter-Satellite Link}
\acrodef{TFRC}{TCP Friendly Rate Control}
\acrodef{RACH}{Random Access CHannel}
\acrodef{LPWAN}{Low-Power Wide Area Network}
\acrodef{VRT}{Variable Rate Technology}
\acrodef{HAP}{High Altitude Platform}
\acrodef{LALE}{Low Altitude Long Endurance}
\acrodef{LASE}{Low Altitude Short Endurance}
\acrodef{MALE}{Medium Altitude Long Endurance}
\acrodef{HALE}{High Altitude Long Endurance}
\acrodef{OTA}{Over The Air}
\acrodef{TLS}{Transport Layer Security}
\acrodef{SSL}{Secure Sockets Layer}
\acrodef{SGD}{Stochastic Gradient Descent}
\acrodef{NN}{Neural Network}

\acrodefplural{EA}[EAs]{Essential Attributes}
\acrodefplural{RSU}[RSUs]{Road Side Units}
\acrodefplural{VANET}[VANETs]{Vehicular Ad-Hoc Networks}
\acrodefplural{OBU}[OBUs]{On-Board Units}
\acrodefplural{UAV}[UAVs]{Unmanned Aerial Vehicles}
\acrodefplural{FANET}[FANETs]{Flying Ad-Hoc Networks}
\acrodefplural{MANET}[MANETs]{Mobile Ad-Hoc Networks}
\acrodefplural{VANET}[VANETs]{Vehicle Ad-Hoc Networks}
\acrodefplural{UAS}[UASs]{Unmanned Aerial Systems}
\acrodefplural{HAP}[HAPs]{High Altitude Platforms}
\acrodefplural{LALE}[LALEs]{Low Altitude Long Endurance}
\acrodefplural{LASE}[LASEs]{Low Altitude Short Endurance}
\acrodefplural{MALE}[MALEs]{Medium Altitude Long Endurance}
\acrodefplural{HALE}[HALEs]{High Altitude Long Endurance}
\acrodefplural{RCST}[RCSTs]{Return Channel Satellite Terminals}
\acrodefplural{VNF}[VNFs]{Virtual Network Functions}

%% file: sections/Introduction.tex
\section{Introduction}
\label{sec:introduction}
The 3D networks are expected to exploit satellite, aerial and terrestrial platforms jointly \cite{bacco2017survey} for improving the radio access in future 6G networks to support the well-known 5G classes of service, i.e., enhanced Mobile Broadband (eMBB), Ultra-Reliable Low Latency Communications (URLLC), massive Machine Type Communications (mMTC), as well as tactile Internet and virtual reality applications\cite{gupta2019tactile}.
However, these 3D networks pose a challenge to the different levels of Quality of Service (QoS) requirements that each network technology can provide. For instance, while the satellite link can provide bandwidth required for an eMBB-type service, it is not guaranteed that it can satisfy the latency constraints expected in 5G for URLLC traffic \cite{bacco2019networking}.
Therefore, service classification and separation are of utmost importance to guarantee the required QoS level to a given traffic flow through the suitable network trunk. Moreover, different access technologies (satellite, aerial or terrestrial) may use data protection techniques \cite{celandroni2011performance, gotta2008experimental, bacco2018tcp}, which may not be suitable for every class of traffic, which introduces in some cases, severe delays or jitter, not matching service class requirements.

Traffic characterization depends on the information found in the client-server data streams. The availability of such information depends on the communication protocols used and the modality with which the data streams are examined \cite{survey}, i.e., data processing techniques play an important role in information retrieval. If the traffic is not encrypted, it is usually possible to perform a complete analysis. In such a scenario, the packet content can be inspected by accessing the packet header and the payload called \ac{DPI}.
Accessing transport protocol information allows header compression/suppression when capacity is scarce, adapt the protocol behaviour to the characteristics of the network bottlenecks (e.g. HTTP acceleration or split-connections), and implement multi-class per-hop behaviour in the absence of another IP signalling. Therefore, a solution is needed to avert losing these benefits when encrypting transport headers \cite{issues} or other HTTP tunnels will be prevalent.
Traditional methods of traffic analysis were based on \ac{DPI} \cite{bujlow2015independent}, which refers to a set of analysis tools aimed at extracting information from the headers, payload and classifying the flows. However, with the increasing number of new applications, which no longer have fixed port numbers that can be queried but adopt random port strategies, the accuracy of
\ac{DPI} are gradually declining. Moreover, since the advent of the encrypted transport headers, as in QUIC traffic \cite{rfc9000} (an encrypted transport protocol proposed by Google), the range of applicability of \ac{DPI} has been progressively fading out, paving way for new form of blind data extraction. 
\ac{ML} is also gaining popularity in many communication, and networking scenarios \cite{saso_14,oceans_19} and recently, in solving traffic classification problems as a viable substitute for \ac{DPI}. \ac{ML} algorithms can be classified into three families: \textit{Supervised}, \textit{Unsupervised} and \textit{Semi-supervised}.
The first family requires labelled training dataset; the second one discover hidden patterns within the dataset without labels; the last one uses a small amount of labelled data and a large amount of unlabelled data during training.

Semi-supervised methods are particularly attractive when the goal is to reduce the cost of obtaining the labelled data. The goal is to retrieve knowledge about the missing labels by evaluating the similarity between unlabeled and labelled data. They typically define data inspection tasks or optimised manual labelling of the unlabeled data to provide adequate information. Even though the semi-supervised techniques do not require a fully labelled data set, the effort required to label some data at the beginning of the training phase can be time consuming. 
To reduce the computational cost of the semi-supervised techniques, dimension reduction methods and federated learning protocols can be used. In the first case, feature selection can be used to reduce the computational and communication costs compared to the elaboration of the native dataset. In the second case, federation allows some of the computational operations to be offloaded to the edge gateways, reducing the computational load for a central server and the number of control messages exchanged.

In this paper, we provide the following contributions:
\begin{itemize}
    \item a federated semi-supervised regression algorithm based on Neural Network suitable for resource constrained devices 
    \item a federated dimensionality reduction procedure based on an information-based feature selection
    \item a comparison of selected state of the art feature selection schemes
    \item a numerical analysis of the performance obtained by using both centralized and federated versions of the regression algorithm
\end{itemize}

The rest of the paper is organized as follows. In Section \ref{sec:relatedworks}, the related work is presented. Section \ref{sec:SoW} presents the general methodology and description of the scenario. The performance evaluation is shown in Section \ref{sec:perf} and Section \ref{sec:conclusions} concludes the paper.

%% file: sections/RelatedWorks.tex
\section{Related Work}
\label{sec:relatedworks}
Internet traffic classification using supervised, unsupervised, and semi-supervised learning has been widely investigated in literature \cite{survey} and several solutions have been studied for traffic monitoring and analysis applications.  These  methods differ in the way they extract features $x_i$, for traffic analysis i.e., the variables that allow outputting the classification $y$. 
Traffic analysis methods fall into three main classes: \textit{Statistics-based methods}, \textit{Correlation-based methods}, and \textit{Fingerprint-based methods}.
\subsubsection{Statistics-based methods}
are based on the representation of traffic flows by statistical features at the traffic flow level, which are used to capture the characteristic patterns of a traffic flow. These features are: statistical metrics related to the number of packets transmitted, the number of bytes transmitted, the time between the arrival of packets, and the packet size of a flow. In \cite{zhang2019stnn},
23, metrics such as maximum length (bytes), minimum length, mean, median, standard deviation, and cumulative length are applied to five objects: Total packets, Forward packets, Backward packets, Handshake packets, and Data transfer packets. In \cite{aceto2018multi}, the statistical features are extracted from a vector of packet lengths, considering three sets of packets: incoming, outgoing, and bidirectional flow packets. In \cite{tong2017accelerating}, the selected features are: Protocol, Source Port Number, and Destination Port Number, which are referred to as classical features; instead, maximum, minimum, mean, and variance of packet size are referred to as statistical features.

\subsubsection{Correlation-based methods} are characterized by incorporating knowledge about the correlation between flows to accomplish the classification task. These methods are often based on the so-called \ac{BoF}, which groups the correlated flows according to a set of heuristics such as the destination IP address, the destination port, and the transport protocol \cite{divakaran2015slic, wang2013internet,wang2011novel}. 
%In \cite{divakaran2015slic} a self-learning classifier called SLIC (Self Learning Intelligent Classifier), capable of performing continuous model updates, is proposed. The SLIC scheme consists of two main components: a classifier based on k-Nearest Neighbours (k-NN) and a decision-maker. %The classifier is trained and used to predict the flow labels, while the decision-maker selects new useful labelled flows to enrich the training set and update the model. 
%The classification method is based on the pre-processing of network traffic into \ac{BoF}, based on the 3-tuple heuristic. Two semi-supervised methods are proposed in \cite{wang2013internet,wang2011novel}, to classify the traffic concerning the communication protocols employed. The role of the previously seen \ac{BoF} is carried by the equivalence set constraints in \cite{wang2013internet,wang2011novel} that are built using the 3-tuple heuristic and indicate that a set of flows are likely to share the same application layer protocol and hence have to be classified equally.
\subsubsection{Fingerprint-based methods} are based on extracting a fingerprint, i.e., a vector or function, possibly a probability density function, capable of summarizing the main characteristics common to a specific traffic class.  %The fingerprinting process starts by dividing data according to the class to which it belongs. Then a fingerprint is extracted per each class. The set of fingerprints together with an opportune metric compose a tool to classify new traffic flows: once a new flow is captured from the network, it is compared to the fingerprints using the chosen metric and it is assigned to the class corresponding to the closest fingerprint. 
Fingerprinting requires a significant amount of data per class to be identified. In \cite{crotti2007traffic}, traffic fingerprinting is used to classify IP flows produced by network applications exchanging data through TCP connections such as HTTP, SMTP, SSH by looking at client-server or server-client flows. 
%After observing several flows from the same protocol, the gathered statistical information is used to build protocol fingerprints to classify an unknown flow. 
A vector of Probability Density Functions (PDF) is estimated from a training set of flows generated by the same protocol to build the protocol fingerprints.
%In \cite{kohout2017network} a method to represent communication patterns compactly is proposed. Communications are abstractly considered like sets of messages exchanged, represented as vectors in a multidimensional space. Each set of messages is treated as observations of a random variable with an unknown probability distribution. The joint distribution of features that describe a flow is represented as a single vector of fixed dimension, the fingerprint. 

In recent years, the proliferation of encrypted traffic has led to an increase of \textit{flow-based methods} that rely on the analysis of statistical or time-serial features using ML. These include Naive Bayes~(NB), Support Vector Machine~(SVM), Random Forest~(RF), and K-Nearest Neighbours~(KNN)~\cite{pacheco,survey2}. 
However, the most tedious task in building a ML model for traffic classification is labelling the data, which requires human intervention, so it is advantageous to use a semi-supervised approach.  %This approach was fine-tuned in~\cite{dcgan} where a deep convolutional generative adversarial network (DGCAN) was used to classify encrypted connections. Their approach provided an accuracy of 89\% when only 10\% of the dataset was labelled.

In~\cite{cnn}, the classifier can detect several classes of services with encrypted traffic with reasonable accuracy. While the initial analysis considers 1400 features, it turns out that these features can be reduced to only three by feature reduction.

%% file: sections/ProblemDefinition.tex
\section{Problem Definition and Methodology}
\label{sec:SoW}

%Since the output of our traffic analysis is a variable with real value and no in a categorical set, i.e. an integer set hence, 
The task we address in this paper is to estimate the percentage of traffic of a particular protocol or class of applications within a data stream that contains many types of traffic. We have applied the regression techniques in \cite{FERNANDEZDELGADO201911} to an edge computing scenario where edge gateways are resource constrained.
In this reference scenario, model training is distributed, i.e., each edge gateway trains the model using its own locally collected traffic. Then, a centralized edge server merges the knowledge obtained from each end gateway and distributes the merged model to all federated end gateways. For this purpose, we developed a regression algorithm based on a neural network
(NN) and combined the regression algorithm with a feature selection technique that is also used in the federation. We then analyzed the accuracy of the federated regression model compared to the centralized model and investigated how feature selection affects model accuracy as well as computational and communication load during model training.
In the following sections, we elaborate on the scheme we developed.
\subsection{Regression Algorithm}
The regression algorithm developed in this work is based on a \ac{NN} with a single hidden layer and uses the ADAM optimizer \cite{kingma2017adam}, an efficient version of
 \ac{SGD}. Unlike  \ac{SGD},
however, ADAM computes individual adaptive learning rates for different parameters from estimates of the first and second moments of the gradients, making it suitable for non-convex optimization problems, even when running on constrained devices, since it is computationally efficient.

\begin{figure}
    \centering
    \begin{subfigure}[b]{0.80\columnwidth}
         \includegraphics[width=\columnwidth]{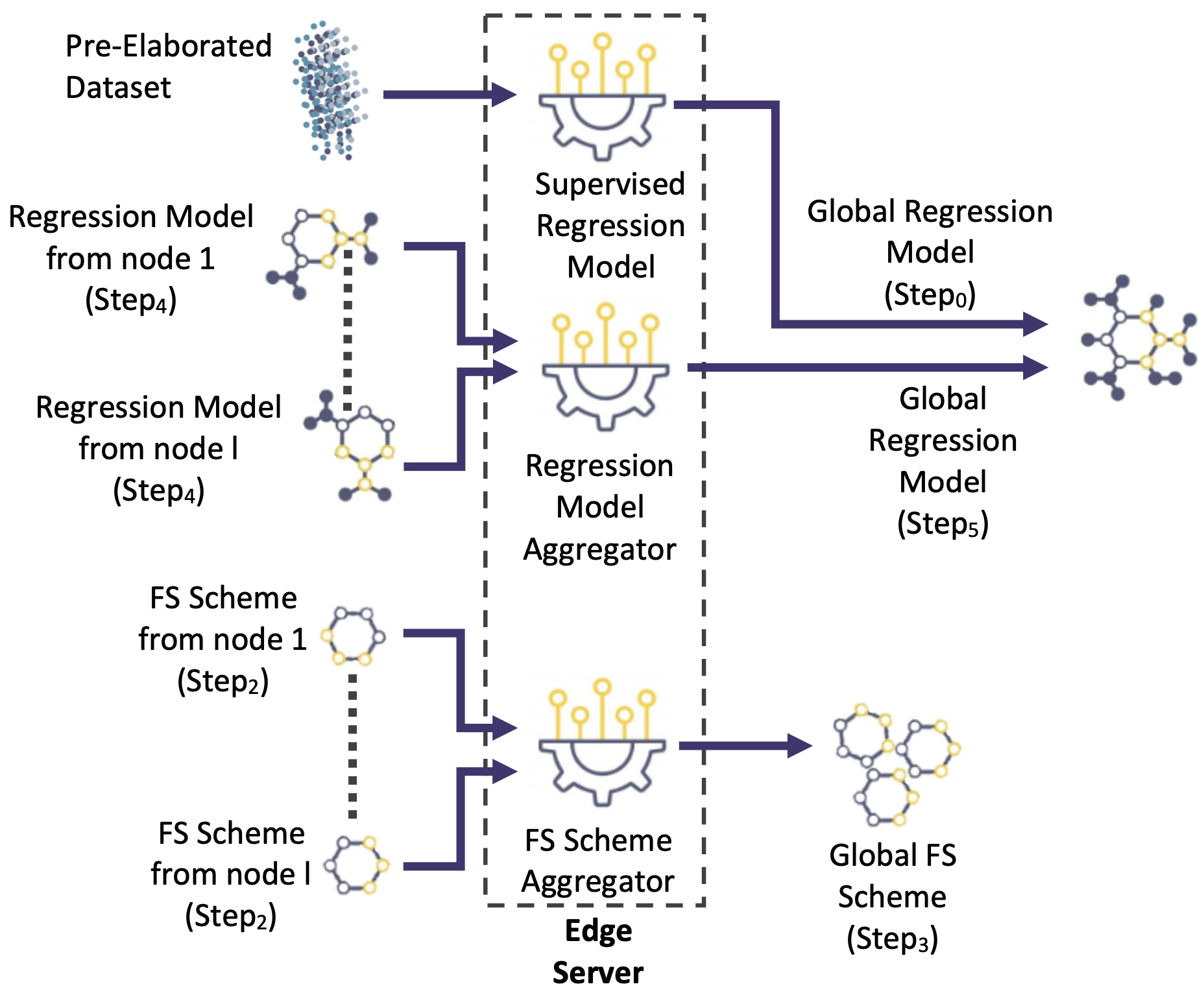}
         \caption{Edge Server}
         \label{fig:1a}
    \end{subfigure}
    %\hfill
    \begin{subfigure}[b]{1\columnwidth}
             \includegraphics[width=\columnwidth]{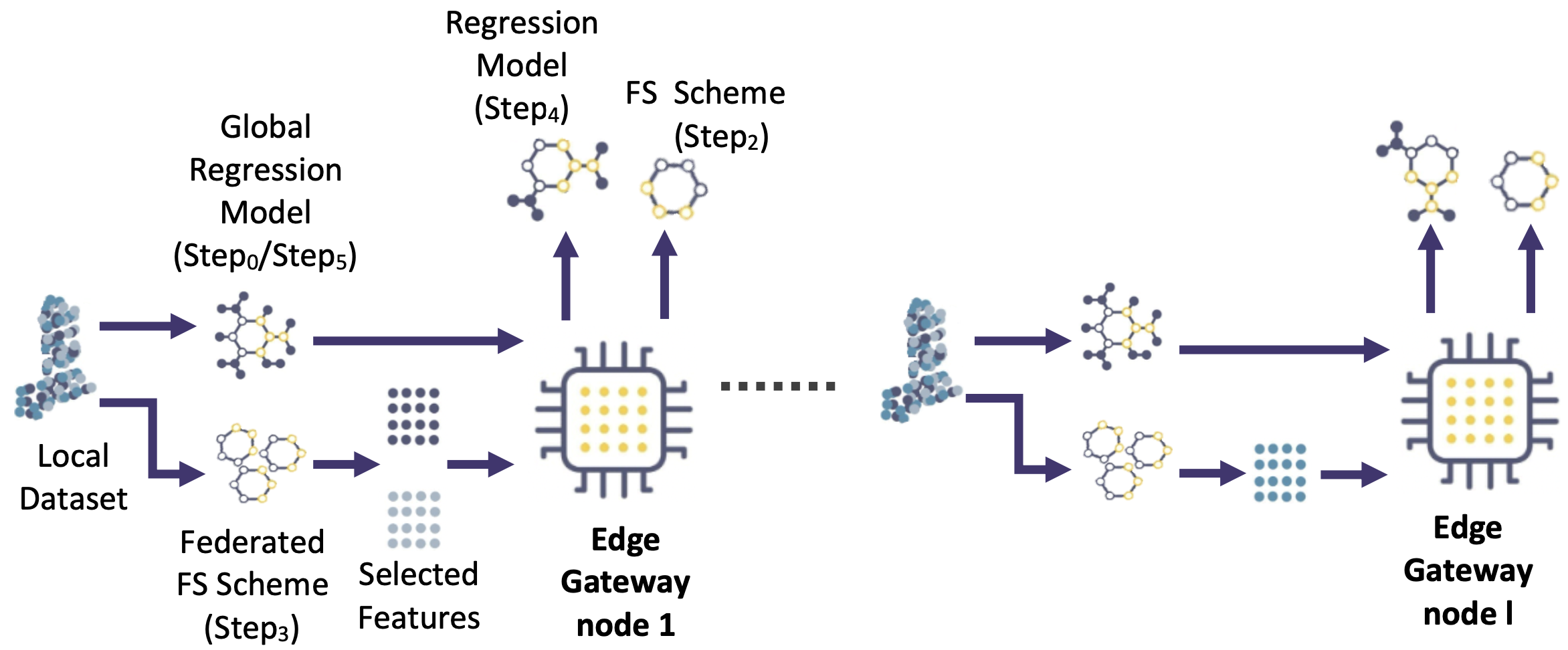}
             \caption{Edge Gateway Node}
             \label{fig:1b}
    \end{subfigure}
    \caption{Federated learning edge architecture}
    \label{fig:1}
\end{figure}

%ADAM algorithm  The main advantages in using this algorithm lie in:  it is straightforward to implement, it is computationally efficient, it can work online, it requires minor memory requirements, it is suitable for non-stationary objective functions, even when their gradient is very noisy or sparse.
 %The algorithm maintains a learning rate for each network weight, separately adapted as learning unfolds. Instead of adapting the parameter learning rates based on the average first moment, Adam also uses the average of the second moments of the gradients. The average used by the algorithm is the exponential moving average of both the gradient and the squared gradient, with the parameters that control the decay rate of the average adapted step by step.\\
%Note that the ADAM algorithm combines the advantages of both algorithms, the Adaptive Gradient Algorithm (AdaGrad) and the Root Mean Square Propagation (RMSProp). The former allows to maintain a per-parameter learning rate improving the performance on problems with sparse gradients, and the last one still maintains per-parameter learning rates that in this algorithm are adapted using the average of current magnitudes of the gradients for the weight.\\
%In the following, we discuss the developed federated procedure used for estimating the percentage of traffic due to either a given protocol or a class of application within a data stream containing many types of data flows through the resolution of a regression problem.
Figure \ref{fig:1}  shows the steps of the proposed procedure; precisely, Fig. \ref{fig:1a} shows the steps performed at the edge server and Fig. \ref{fig:1b} shows the steps performed at the edge gateways.
$\mathbf{Step_0}$ \quad The edge server trains the initial regression model
in a supervised way using a pre-elaborated dataset, and propagates the trained model to the edge gateways.
$\mathbf{Step_1}$ \quad Each edge gateway acquires unlabelled local data and performs
soft data labelling by using the global model.\\
$\mathbf{Step_2}$ \quad Each edge gateway uses its soft labels to perform feature selection, and sends the \ac{PMF} of the selected features to edge server.\\
$\mathbf{Step_3}$ \quad The edge server aggregates all these selected features \ac{PMF}, and send it back to the edge gateways the federated feature selection scheme.\\ 
$\mathbf{Step_4}$ \quad Each edge gateways trains its local models using the restricted set of features evaluated through the federated scheme. The trained model is then transmitted toward the edge server for the federation\\
$\mathbf{Step_5}$ \quad The edge server averages the received models from edge gateways and merges this averaged model with the global regression model available. The edge server propagates the new global regression model toward the edge gateways. \smallskip\\
The process continues from $\mathbf{Step_1}$ until a desired accuracy is achieved. The following section explains in detail the feature selection algorithm.
\subsection{Feature Selection Algorithm}
The \ac{FS} algorithm used in this paper is developed in both centralised and federated fashion, inspired by \cite{cassara2021federated}, and can be formulated as follows: 
\begin{Def}[\ac{FS} Problem]
Given the input data matrix $\mathbf{X}$ composed by $n$ samples of $m$ features ($\mathbf{X} \in \mathbb{R}^{n \times m}$), and the target labels' vector $\mathbf{y} \in \mathbb{R}^{n}$, the feature selection problem is to find a $k$-dimensional subset $\mathbf{K} \subseteq \mathbf{X}$ with $k \leq m$, by which we can characterize $\mathbf{y}$.
\end{Def}
The algorithm used in this work is based on the measure of \ac{MI} that measures the amount of information obtained about the class label through the set of selected features. The  \ac{MI} is related to the entropy $\mathbf{H}(\cdot)$ that measures the uncertainty of a random variable, as shown in the following equation \cite{eliece77, cover91}.
\begin{equation} 
  \label{MI_3}
  \mathbf{I}(\mathbf{U};\mathbf{y})=\mathbf{H}(\mathbf{y})-\mathbf{H}(\mathbf{y}|\mathbf{U}),
\end{equation}
In the previous equation $\mathbf{U}=\{\mathbf{x}_1 \cdots \mathbf{x}_k \; | \; k\leq m\} \subseteq \mathbf{X}  $ is the subset of selected features, and $\mathbf{H}(\mathbf{y}|\mathbf{U})$ is the conditional entropy that measures the amount of information needed to describe $\mathbf{y}$, conditioned by the information carried by $\mathbf{U}$. In brief, the $ \mathbf{I} (\mathbf{U};\mathbf{y})$ represents the dependence between $\mathbf{U}$ and $\mathbf{y}$, i.e., the greater is the value $\mathbf{I}$, the greater is the information carried by $\mathbf{U}$ on $\mathbf{y}$. 
The features selected in $\mathbf{U}$, also known as \acp{EA}, provide the maximum value for equation (\ref{MI_3}).

In \cite{eliece77, cover91} authors prove that the feature selection problem can be solved as an optimization problem as given in equation (\ref{Native Optimization Porblem}), 
 \begin{gather} 
\label{Native Optimization Porblem}
\displaystyle \mbox{arg}\max_{\mathbf{U}} \mathbf{I}(\mathbf{U};\mathbf{y}) \\
\nonumber
\mathbf{U}=\{\mathbf{x}_1 \cdots \mathbf{x}_k \; | \; k\leq m\} \subseteq \mathbf{X}
\end{gather}%
by assuming independent features. They also provide the incremental version of the previous algorithm, as shown in equation (\ref{Incremental Optimization Porblem}).
\begin{gather} 
\label{Incremental Optimization Porblem}
\displaystyle \mbox{arg}\max_{\mathbf{x}_j \in \mathbf{X}\setminus\mathbf{U}} \mathbf{I}(\mathbf{x}_j;\mathbf{y}|\mathbf{U}), \\
\nonumber
\mathbf{U}=\{\mathbf{x}_1 \cdots \mathbf{x}_{k-1} \; | \; k\leq m\} \subseteq \mathbf{X}.
\end{gather}
By using Cross-Entropy-based \cite{rubinstein04} feature selection algorithm, we can provide the solution to equation (\ref{Native Optimization Porblem}) by selecting a set of \ac{EA}s at a time, instead of selecting a single feature at a time. The solution provided by the Cross-Entropy-based algorithm is the distribution probability $\mathbf{p}=[p_1,\cdots p_m]$ of selecting features in $\mathbf{U}$ to get the maximum for the problem in equation (\ref{Native Optimization Porblem}).

The \ac{FFS} procedure adopted in this paper exploits the Bayesian's theorem to merge the local distribution probability $\mathbf{p}$ into the global one.
Formally, we assume that each node acquires several \textit{i.i.d.} records $n^{l}$  to address the optimization problem (\ref{Native Optimization Porblem}), and that the nodes share the same set of features $\mathbf{X}$.
 The global probability $\mathbf{p}^{\mathbf{G}}$ used for the \ac{FS} can be written as follows:
\begin{equation}
\label{eq:joint_probability}
    \mathbf{p}^{G}=\sum_l \mathbf{p}^{l} q^{l},
\end{equation}
 where $\mathbf{p}^{(l)}$ is the probability distribution of selecting the features at the node $l$, and $q^{(l)}$ is the weight of this distribution probability. As shown in equation (\ref{eq:probability_data_from_node}), the weight 
is proportional to the size of its local dataset compared to the whole amount of data present in the system. In this way, we can contrast situations where local datasets are heterogeneous w.r.t. the size.
\begin{equation}
    \label{eq:probability_data_from_node}
    q^{l}=\frac{n^{l}}{\displaystyle\sum_l n^{l}}
\end{equation}
In the next section we provide the performance analysis of the procedure discussed in this section.

%% file: sections/PerformanceEvaluation.tex
\section{Numerical Analysis}
\label{sec:perf}
The dataset used to analyse the performance of the developed semi-supervised method is the QUIC traffic dataset \cite{datasetQUIC}. The dataset contains flow packets based on QUIC and non-QUIC traffic generated from five different Google services: Google Drive, Google Docs, Google Music, Google Search, and YouTube. The authors developed scripts using Selenium WebDriver and AutoIt tools to mimic human behaviour during data collection. This approach enables the acquisition of data sets with more than $20\cdot 10^6$ records without much human effort.
%The data are recorded in many directories, one for each service, containing many files deriving from different days of acquisition. 
The records in the files contain three fields: Unix timestamp, the acquisition time and packet length

We merged all the files by synchronizing them with the Unix timestamps. This gave us a single file containing all the records for QUIC and non-QUIC traffic for the five different classes of services as mentioned above. We also developed Python-based scripts to extract the following features: Number of QUIC packets within a sampling window of size $1$ second, $25$-th, $50$-th, $75$-th, and $90$-th percentiles of packet arrival time, $25$-th, $50$-th, $75$-th, and $90$-th percentiles of packet sizes. For each record of extracted features, we evaluated two class labels: \textit{Class Label QUIC}- the percentage of QUIC-based traffic and \textit{Class Label Service}- the percentage of traffic due to a class service within the sample window considered. We split the dataset with the extracted features into two sets: The first contains the $20\%$ of samples used for supervised training of the regression model, as specified in the $\mathbf{Step_0}$ of the procedure. The second contains the $80\%$ of samples distributed over a set of $10$ end devices to perform the local unsupervised training of the regression model, as specified in the procedure from $\mathbf{Step_1}$ to $\mathbf{Step_5}$.

We analyzed the performance of the proposed cross-entropy-based (CE) algorithm for feature selection from the QUIC dataset. We compared the performance of CE in detecting the proportion of QUIC traffic versus non-QUIC traffic in a sample of traffic outages with the performance of other information-based feature selection algorithms such as mRMR \cite{peng2005feature}, CMIM \cite{meyer2008information}, and DSR \cite{fleuret2004fast}, which provide the solution to the optimization problem in equation (\ref{Incremental Optimization Porblem}). We also compared the cross-entropy based algorithm with the analysis of variance technique. This technique, known as ANOVA \cite{heiberger2009one}, determines the variance of all features and divides them into systematic and random factors, where random factors have no effect on learning because the variance of these features is zero. The higher the F-score, the greater the variance between the means of the two populations. We assume that features with zero variance do not add information by considering the relationship between the target variable and the feature vectors. In our case, ANOVA is used to compare the mean values of the features and determine a set of features that efficiently contribute to the classification of traffic.
Table \ref{table:feature_set} shows the results of the feature selection scheme obtained with the methods discussed above.
\begin{table}[ht]
\centering
\begin{tabular}{m{0.25cm}>{\centering} m{1cm}|>{\centering} m{0.75cm}>{\centering} m{1cm}>{\centering} m{0.75cm}>{\centering} m{0.75cm}c} 
\hline
\multicolumn{2}{c|}{ }&\multicolumn{5}{c}{\textbf{Features Selection Scheme}}\\\hline
\multicolumn{2}{c|}{\textbf{Feature}} & \textbf{CE} & \textbf{ANOVA} & \textbf{CMIM} & \textbf{DISR} & \textbf{mRMR}\\\hline
$1)$  & $\overline{N}$     & 0 & 0 & 0 & 1 & 1\\\hline
$2)$  & $\Delta T_{25-th}$ & 0 & 1 & 0 & 0 & 1\\\hline
$3)$  & $\Delta T_{50-th}$ & 1 & 0 & 1 & 1 & 0\\\hline
$4)$  & $\Delta T_{75-th}$ & 0 & 1 & 0 & 0 & 0\\\hline
$5)$  & $\Delta T_{90-th}$ & 0 & 0 & 0 & 0 & 0\\\hline
$6)$  & $Ln_{25-th}$       & 1 & 0 & 1 & 0 & 0\\\hline
$7)$  & $Ln_{50-th}$       & 1 & 1 & 1 & 1 & 1\\\hline
$8)$  & $Ln_{75-th}$       & 1 & 1 & 1 & 1 & 1\\\hline
$9)$  & $Ln_{90-th}$       & 1 & 1 & 1 & 1 & 1\\\hline
\end{tabular}
\caption{Extracted Features Set}
\label{table:feature_set}
\end{table}
Most feature selection methods provide a ranking of the features analyzed. Instead, our method automatically returns the minimum set of features, which in this case is $5$ and guarantees the best estimate for the regression model. For this reason, we select the first five features of each method. Another advantage of the cross-entropy based algorithm is that we can implement a federated version. To our knowledge, this is the first federated feature selection method. Table \ref{tab:main_results} shows the comparison of the performance obtained with the regression model in identifying the percentage of QUIC traffic in the received data stream, taking the subsets of features selected with the various methods discussed previously. The cross-entropy and CMIM algorithms provide a performance degradation of the regression model of just under $1\%$ \textit{w.r.t.} without feature selection; instead, we obtain a degradation of almost $5\%$ with ANOVA and almost $10\%$ with both DSR and mRMR, compared to the performance obtained with the full set of features. While we obtain the same performance with the cross-entropy-based and CMIM algorithms, the former allows us to automatically compute the minimum number of features and can also be implemented in a federated version.
\begin{table}[]
    \centering
    \begin{tabular}{llll}
    \toprule
    \multicolumn{2}{c}{Method} & RMSE& Selected Feat.\\
    & &  & \# \\
    \midrule
    
     No Feature selection & -- & $0.0121$ & 9 (All)\\\\
    \multirow{6}{*}{Centralised FS} & CE & $0.0122$ & 5\\
    & ANOVA & $0.0127$ & top 5\\
    & CMIM & $0.0123$ & top 5\\
    & DSR & $0.0133$ & top 5\\
    & mRMR & $0.0133$ & top 5\\\hline
  \end{tabular}
    \caption{Supervised Feature selection. Performance of regression model trained using the features selected by each method. }
    \label{tab:main_results}
\end{table}
In Table (\ref{tab:main_results_2}), we provide the performance analysis in terms of control traffic (measured in MBs) when both FS and the regression model are federated. More specifically, we compare the performance of the procedure used to train the regression model when this is done without model federation and feature selection, or Centralized Regression (CR), with federation of models, or Federated Regression (FR), and with federation of models trained over the selected set of features, or Restricted Federated Regression (RFR).
\begin{table}[hbt!!]
    \centering
\begin{tabular}{m{1cm} m{1.5cm} m{1.8cm} m{0.8cm}}
    \toprule
Procedure & Avrg. Traffic & Conv. Rounds & RMSE\\
          & (Fed./F.S.)     & (Fed./F.S)  &     \\
    \midrule
  CR          & $5.8$          & -     &$0.0122$\\
  FR            &$0.44$/-        & $8$/-&$0.0152$\\
  RFR &$0.315$/$0.108$ & $10$/$12$&$0.0219$\\ \hline
  \end{tabular}
    \caption{Performance comparison among the regression procedures: centralized, federated and restricted federated. }
    \label{tab:main_results_2}
\end{table}
The values in the column \textit{Avrg. Traffic} show the volume of control traffic generated by all nodes during the learning procedure to train the regression model. Centralized regression generates the highest load because all nodes use regression model estimation using a neural network with an input layer based on all available data sets in each learning round. The federated regression involves lighter regression models than the previous one because the neural network input layer is based on a smaller data set, namely the data set available at the node. In this case, all control traffic is due to the federated method. Finally, the Restricted Federated Regression procedure generates the least control traffic, since it uses the local dataset containing only the records of the selected features. Note that in this last case $0.108$ MB of the $0.315$ MB control traffic is due to the federated feature selection procedure, which involves data exchange between the edge server and the end devices.

The values in the \textit{Conv. rounds} column indicate the number of communication rounds required by the federated procedure to merge the local models into the global model. We assume that convergence for the global model is achieved when the difference between the weights of two consecutive rounds of learning is on average less than $1\%$. The average number of rounds of communication required for convergence is also shown during a learning round for the Restricted Federated Regression procedure.

Finally, the values in the \textit{RMSE} column show the values of RMSE evaluated between the ground truth and the trained regression model trained with the procedures discussed so far. The lowest RMSE is obtained with centralized regression, since we use the entire dataset to feed the neural network. Note that the RMSE increases when the amount of information used as input to the neural network is reduced. Note that the degradation of the RMSE is slower than the growth of the control traffic. For example, the RMSE of the Restricted Federated Regression is $\sim 1.8\times$ the RMSE of the Centralized Regression, but the control traffic of the Centralized Regression is more than $50\ times$ the control traffic of the Restricted Federated Regression.